\documentclass[conference,10pt]{IEEEtran}
\IEEEoverridecommandlockouts

\usepackage{cite}
\usepackage{amsmath,amssymb,amsfonts}
\usepackage{algorithmic}
\usepackage{graphicx}
\usepackage{textcomp}
\usepackage{xcolor}
\usepackage{amsmath}
\usepackage{subcaption}
\usepackage{multirow}
\usepackage{balance}
\usepackage{ctable}
\usepackage{soul}

\usepackage{threeparttable}
\def\BibTeX{{\rm B\kern-.05em{\sc i\kern-.025em b}\kern-.08em
    T\kern-.1667em\lower.7ex\hbox{E}\kern-.125emX}}
\begin{document}

\title{LLMs Explain't: A Post-Mortem on Semantic Interpretability in Transformer Models}


\author{\IEEEauthorblockN{
        Alhassan Abdelhalim\IEEEauthorrefmark{1},
        Janick Edinger\IEEEauthorrefmark{1}, 
        Sören Laue\IEEEauthorrefmark{2} and
        Michaela Regneri \IEEEauthorrefmark{2}
        }
    \IEEEauthorblockA{\IEEEauthorrefmark{1} Distributed Operating Systems Group, Department of Informatics, Universität Hamburg, Germany}
    \IEEEauthorblockA{\IEEEauthorrefmark{2} Machine Learning Group, Department of Informatics, Universität Hamburg, Germany}

Email: \{alhassan.abdelhalim, janick.edinger, soeren.laue, michaela.regneri\}@uni-hamburg.de
    
}

\maketitle

\begin{abstract}
Large Language Models (LLMs) are becoming increasingly popular in pervasive computing due to their versatility and strong performance. However, despite their ubiquitous use, the exact mechanisms underlying their outstanding performance remain unclear. Different methods for LLM explainability exist, and many are, as a method, not fully understood themselves. We started with the question of how linguistic abstraction emerges in LLMs, aiming to detect it across different LLM modules (attention heads and input embeddings). For this, we used methods well-established in the literature: (1) probing for token-level relational structures, and (2) feature-mapping using embeddings as carriers of human-interpretable properties.
 Both attempts failed for different methodological reasons: Attention-based explanations collapsed once we tested the core assumption that later-layer representations still correspond to tokens. Property-inference methods applied to embeddings also failed because their high predictive scores were driven by methodological artifacts and dataset structure rather than meaningful semantic knowledge. These failures matter because both techniques are widely treated as evidence for what LLMs supposedly understand, yet our results show such conclusions are unwarranted. These limitations are particularly relevant in pervasive and distributed computing settings where LLMs are deployed as system components and interpretability methods are relied upon for debugging, compression, and explaining models.

\end{abstract}


\section{Introduction}

Large language models (LLMs) are now central to research and applications across Natural Language Processing and other fields of computer science. As their capabilities grow, so does the pressure to explain how these systems represent linguistic structure, acquire conceptual knowledge, and integrate information across layers. In Pervasive Computing and Edge AI, several methods such as Pruning \cite{ma2023llm}, Quantization \cite{cai2024self}, and Knowledge Distillation \cite{hasan2025optimclm,hu2025efficient}, are used to compress LLMs in order to make them executable on small Edge devices.  However, those compression methods suffer from the loss of the little explainability we can establish for larger models \cite{yeom2021pruning,alharbi2021learning,kdir23,lee2025advancing}. In pervasive and distributed computing, language models are rarely used in isolation. Instead, they are embedded into larger systems and often under strict resource constraints. In such settings, interpretability methods are considered practical tools to guide deployment decisions, model compression, debugging, and failure diagnosis. If misleading or overconfident explanations are presented, they can lead to incorrect system-level conclusions. For this reason, robust interpretability techniques are needed for LLMs to bridge the gap between model compression and explainability preservation. A broad methodological ecosystem for explaining LLMs has emerged, such as attention-head interpretation to probing \cite{belinkov-2022-probing}, feature decoding \cite{fagarasan-etal-2015-distributional,chersoni-etal-2021-decoding}, and geometry-based analyses \cite{ethayarajh-2019-contextual}. Despite their popularity, the explanatory power of those methods remains largely unproven. Our own work attempted to reuse established techniques for LLM explainability, but they proved to lack methodological rigor despite their intuitively compelling approaches.

We set out with an optimistic hypothesis: that established interpretability pipelines such as probing, feature-mapping, attention analysis, and controlled experiments would reveal internal semantic structure in LLMs. This hypothesis was well aligned with the existing literature, which regularly reports that attention patterns reflect linguistic relations such as coreference, dependency heads, and taxonomies \cite{1361981470578490112}, and that embedding vectors implicitly encode human-interpretable semantic or perceptual properties \cite{abnar-etal-2018-experiential,chersoni-etal-2021-decoding}. Prior work further argues that linear or shallow nonlinear mappings can decode conceptual features from hidden states or type-level embeddings, and often treats high predictive performance in such mapping tasks as evidence that the corresponding semantic knowledge is present in the model \cite{PINE}. 

We started with the research question on how linguistic abstraction is represented in LLMs. Abstraction is a core mechanism for language and learning: knowing, e.g., what a "dog" is and eats will give you a good idea of how to deal with any instance of dog in real life, no matter what its concrete breed is. LLMs exhibit advanced language capabilities that suggest they can perform this kind of abstraction, but to date, it is unclear how they achieve it. 

Motivated by these promises, we applied two established interpretability methods: attention-based relational explanations and property inference from embeddings. Our aim was not to overturn prior work, but to replicate its claims and apply it to our own scenario using controlled, transparent experiments. Instead, both pipelines failed as soon as the underlying assumptions that underpinned these methods were systematically tested.
The attention-based analyses failed because the token-level structure they presuppose simply does not persist—representations in deeper layers no longer correspond cleanly to individual tokens, making any inferred “relations” artifacts of residual mixing rather than genuine model-internal structure. The embedding–feature inference pipeline failed for the opposite reason: rather than revealing semantic properties encoded in embeddings, its high predictive performance was driven by dataset geometry and methodological upper bounds, not by meaningful information overlap.

\section{Related Work}

Efforts to explain large language models typically fall into two broad categories: 
(i) \emph{attention-based analyses}, which treat attention weights as indicators of token-to-token relations, 
and (ii) \emph{embedding-based property inference}, which maps internal representations onto interpretable semantic features. 
Both families of methods aim to extract mechanistic insight from models whose internal states are high-dimensional, nonlinear, and deeply entangled. 
Despite their widespread adoption, the empirical and theoretical assumptions underlying these approaches have been examined only partially. 
In this section, we summarize the prevailing narratives behind these two interpretability paradigms and highlight the implicit assumptions 
that our experiments directly stress-test.

\subsection{Attention-Based Explanations}

A large body of work interprets transformer attention heads as encoding linguistic or relational structure between tokens (e.g., ``head $h$ attends from \textit{Labradoodle} to \textit{dog}, so it captures a hypernym relation'') \cite{1361981470578490112}. 
This line of research has produced influential tools for visualizing attention matrices \cite{vig-2019-multiscale}, 
along with claims that specific heads implement syntactic dependencies or semantic associations \cite{kovaleva-etal-2019-revealing}.

Two implicit assumptions underpin these interpretations. 
First is the token continuity assumption, which is the idea that the representation at position $i$ in layer $L$ 
still corresponds to the same lexical token that occupied that position at the input \cite{NEURIPS2022_6f1d43d5}. 
Second is the attention interpretability assumption, which is that attention weights can be read off as indicators of relational importance or information flow \cite{jain-wallace-2019-attention}.

Recent analyses challenge both assumptions. 
Work on representational flow in transformers has shown that residual connections, MLP blocks, and multi-head mixing 
rapidly dissolve token-specific identity as depth increases \cite{elhage2021mathematical}, 
making it unclear whether later-layer vectors correspond to individual words in any stable sense. Previous work on linguistic relations has shown that a token's position in the sequence seemingly contributes more than its actual meaning to the embedding representations - suggesting that the position of the token in question cannot be determined \cite{mickus-etal-2020-mean}.
Moreover, prior studies have noted that attention visualizations often exhibit structured, linguistically plausible patterns 
even when the underlying computation is not relational, creating a form of \emph{visualization fallacy} \cite{wiegreffe-pinter-2019-attention}. 
These concerns call into question whether surface-level attention maps can be treated as mechanistic evidence for token-level reasoning.

\subsection{Embedding-Based Property Inference}

A second major line of interpretability work aims to explain what embeddings ``know'' by predicting human-interpretable 
semantic features from them. This approach, often termed \emph{property inference} \cite{rosenfeld2023analysis}, assumes that if a predictive 
model can map embeddings to a curated set of semantic attributes, then those attributes must be implicitly encoded in the vector space. 
Such studies have employed so-called feature norms, which lists of psychologically relevant properties of words. Those norms can contain semantic knowledge \cite{mcrae2005semantic}, perceptual or cognitive properties \cite{binder2016toward}, and neural activation patterns from functional magnetic resonance imaging (fMRI) targets for interpreting embedding content \cite{mitchell2008predicting}.

The standard pipeline trains a regression model, typically Partial Least Squares Regression (PLSR) or a small feedforward neural network to approximate a mapping $f : \mathbf{X} \mapsto \mathbf{Y}$, where $\mathbf{X}$ represents word or concept embeddings 
and $\mathbf{Y}$ represents interpretable features \cite{rosenfeld2023analysis}. 
High predictive accuracy is then taken as evidence that the embeddings encode the corresponding properties.

This methodology bases on a central prediction-as-explanation assumption, taking successful prediction of properties to indicate that the model has internalized those features as emergent knowledge. Recent analyses challenge this assumption by separating predictive success from explanatory validity. In particular, prior work has shown that both PLSR and small neural networks effectively behave as reduced-rank regression models \cite{izenman1975reduced}, 
making them highly sensitive to dataset geometry and sparsity rather than to semantic content. 

Our own work has shown that high scores can arise even when target features are shuffled, corrupted, or replaced with nonsensical values, and that predictive performance is often governed by methodological upper bounds imposed by sparsity and distributional structure rather than by actual information overlap between embeddings and features \cite{41224}. 

This motivates a more systematic evaluation of the conditions under which these methods produce interpretable conclusions, 
a direction our experiments pursue by testing the robustness of property inference to controlled perturbations and alternative explanations.

\subsection{Assumptions Under Test}

We directly test the previous assumptions. For attention, we examine whether relational interpretations remain coherent once representational flow is traced through residual 
streams and nonlinear transformations. For embeddings, we evaluate whether property inference continues to yield meaningful conclusions once confounding factors such as 
sparsity, upper bounds, and geometric similarity are controlled.

\section{Methodology}

To ensure that our findings reflect methodological limitations and not implementation errors, we designed each experimental pipeline to closely replicate the standard practice in the literature, including established preprocessing steps, model choices, hyperparameter tuning procedures, and commonly used visual or statistical analysis tools. Furthermore, for each approach, we introduced multiple sanity checks and ablation experiments to validate the pipeline correctness before assessing the validity of its explanatory assumptions. In this section, we detail the methodological setup for both explanation pipelines: (1) attention-based relational interpretation and (2) embedding-based semantic property inference. Each subsection follows the same structure: we state the intended hypothesis, replicate the standard pipeline used in prior work, describe the models and datasets, outline the evaluation and control experiments, and clarify the methodological foundations on which the later negative results rest.

\subsection{Attention-Based Explainability}

\subsubsection{Intended Hypothesis and Assumptions}

Attention mechanisms are commonly interpreted as windows into a model's internal reasoning. Numerous studies treat
attention weights as indicators of token--token influence \cite{NIPS2017_3f5ee243} and many visual analysis tools reinforce this assumption by depicting attention patterns as semantic or syntactic dependencies. The prevailing hypothesis is that if a model assigns high attention from token $i$ to token $j$, then token $j$ exerts a meaningful influence on the representation or prediction associated with $i$. In our case, we wanted to study abstraction relations within a sentence, e.g., the attention weights between "dog" and "Labradoodle" in each attention head. In this implementation, there are two central implicit assumptions:

\begin{enumerate}
    \item \textbf{Token identity persists across layers:}
    The hidden representation at each position continues
    to correspond to a specific input token, even in deeper
    layers.
    \item \textbf{Attention reflects information flow:}
    attention distributions encode the flow of meaningful
    content or reasoning steps through the network.
\end{enumerate}

Our methodological goal was to reproduce this standard
interpretation pipeline as faithfully as possible and evaluate
whether these assumptions hold under systematic scrutiny.

\subsubsection{Models and Data}

We conducted all experiments using transformer-based
language models with multi-head self-attention
\cite{NIPS2017_3f5ee243}. For a broad
evaluation, we extracted attention weights from all layers and
heads across a range of typical inputs, including short
sentences, syntactic minimal pairs, and simple relational
prompts. Inputs were chosen to match the settings commonly
used in attention-interpretability research. 
For each tokenized input sequence, we computed the scaled
dot-product attention matrices for every head. Attention maps
were collected both in raw form and after standard aggregation
procedures (averaging across heads, layers, or both), mirroring
prior visualization practice.

\subsubsection{Standard Pipeline and Visual Analysis}

Following established methodology, we implemented the
canonical attention-inspection workflow:

\begin{itemize}
    \item extracting attention maps from each layer and head,
    \item generating heatmaps and dependency-style arc
    diagrams, 
    \item selecting high-attention edges and tracing them as
    hypothesized token-level relations,
    \item aggregating attention patterns across layers to
    identify ``global'' dependencies, and
    \item comparing these emergent patterns with expected
    linguistic or semantic relations.
\end{itemize}

\begin{figure}[h!]
    \centering
    \begin{subfigure}[b]{0.23\textwidth}
        \centering
        \includegraphics[width=\textwidth]{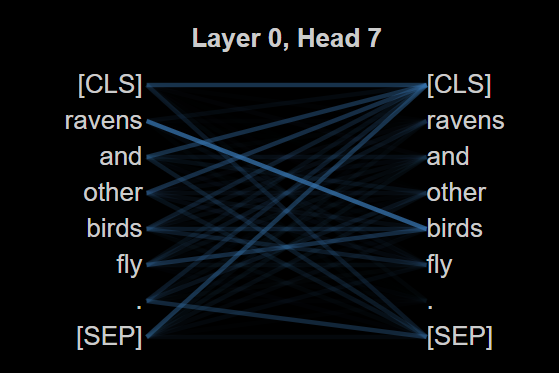}
    \end{subfigure}
    \begin{subfigure}[b]{0.23\textwidth}
        \centering
        \includegraphics[width=\textwidth]{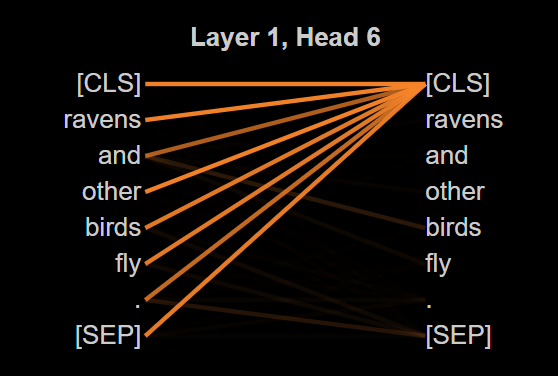}

    \end{subfigure}

     \vspace{0.5em} 
    \begin{subfigure}[b]{0.23\textwidth}
        \centering
        \includegraphics[width=\textwidth]{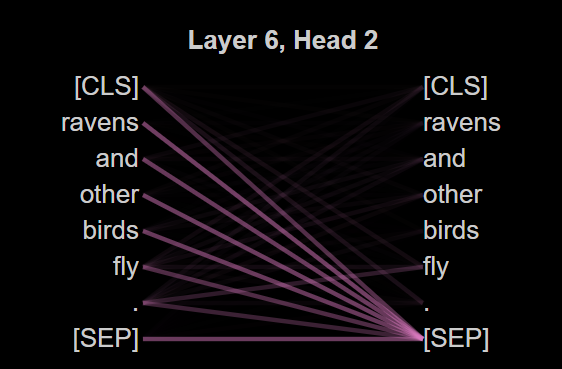}

    \end{subfigure}
    \begin{subfigure}[b]{0.23\textwidth}
        \centering
        \includegraphics[width=\textwidth]{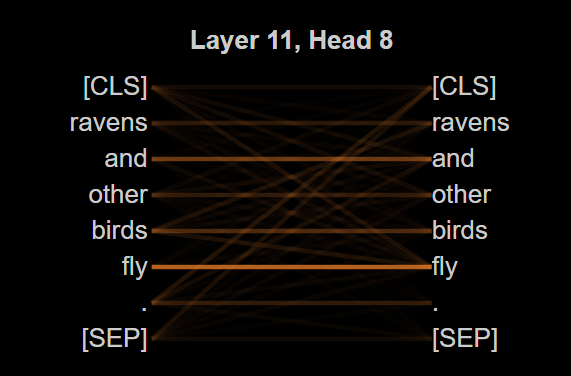}

    \end{subfigure}

    \caption{An exemplary visualization of attention head activations in BERT \cite{devlin-etal-2019-bert}, created with BertViz \cite{1361981470578490112}. We argue that the assumption that words have the same embedding positions across layers is not defensible in the inner layers.}
\end{figure}

\subsection{Property Inference From Embeddings}

\subsubsection{Intended Hypothesis and Assumptions}

Property inference is one of the most widely used approaches
to explain what information is encoded in embedding spaces.
In this paradigm, a predictive model is trained to map word
embeddings onto human-interpretable semantic features such
as perceptual properties, taxonomic categories, or conceptual
attributes. The common assumption, explicitly or implicitly stated in
numerous prior studies, which states that if a model can successfully predict semantic features from an embedding vector, then the embedding must encode the corresponding semantic knowledge.

This assumption underlies multiple claims in the literature
regarding emergent meaning in contextualized and static
embeddings. Our methodological aim was to reproduce this
pipeline faithfully, using standard datasets, standard mapping
methods, and standard evaluation metrics, and then test the
validity of this assumption under controlled ablations.

\subsubsection{Datasets and Feature Norms}

We used three commonly adopted feature-norm datasets, namely McRae \cite{mcrae2005semantic}, Buchanan \cite{buchanan2019english} and Binder \cite{binder2016toward}. Table~\ref{tab:feature_norms} summarizes their key properties.

\begin{table}[h]
\centering
\begin{tabular}{lccc}
\toprule
\textbf{Norm} & \textbf{Type} & \textbf{\# Concepts} & \textbf{\# Features} \\
\midrule
McRae et al.\ (2005) & Categorical & 541 & 2526 \\
Buchanan et al.\ (2019) & Categorical & 4436 & 3981 \\
Binder et al.\ (2016) & Continuous & 535 & 62 \\
\bottomrule
\end{tabular}
\caption{Overview of the feature-norm datasets.}
\label{tab:feature_norms}
\end{table}

For the categorical norms (McRae and Buchanan) and to ensure comparability, we followed
the standard preprocessing paradigm of converting concept-feature
pairs into binary or frequency-weighted matrices
\cite{mcrae2005semantic, buchanan2019english}.

\subsubsection{Embedding Extraction}

Following prior work,
we used type-level embeddings obtained from the 0th layer of
a pretrained transformer model. For multi-token words,
we averaged subword embeddings into a single concept-level
vector.

\subsubsection{Mapping Methods}

We followed the two mapping techniques most widely used
for decoding semantic properties from embeddings:

\begin{itemize}
    \item \textbf{Partial Least Squares Regression (PLSR)}:
    a linear mapping that projects input and output into a
    shared low-dimensional latent space optimized for
    covariance.

    \item \textbf{Feedforward Neural Networks (FFNNs)}:
    a single-hidden-layer MLP with tanh activation, used as
    a nonlinear alternative.
\end{itemize}

Both methods were implemented using the same hyperparameter
settings reported in prior studies.

\subsubsection{Hyperparameter Selection and Overfitting Controls}

A critical methodological issue noted in prior work is that the performance of both PLSR and FFNN regressors depends sensitively on the choice of the latent dimensionality $k$ \cite{Mikolov2013EfficientEO}. To ensure a fair and stable comparison, we selected $k$ by monitoring train–validation MSE curves to detect overfitting, identifying the elbow point that minimized validation error, and then applying the resulting value consistently across both PLSR and FFNN models. This procedure ensures that both mapping methods operate within their optimal capacity and avoids artificially inflated performance due to over-parameterization, which earlier work has demonstrated to be a major source of misleadingly high scores in embedding–feature mapping \cite{PINE}.

\subsubsection{Evaluation Metrics}

We replicated the standard evaluation measures from the
literature \cite{mcrae2005semantic,buchanan2019english,binder2016toward,rosenfeld2023analysis}:

\begin{itemize}
    \item \textbf{F1@10} for categorical norms (McRae,
    Buchanan)
    \item \textbf{Spearman's $\rho$} for continuous norms
    (Binder)
    \item \textbf{Neighborhood Accuracy@10} as a geometric
    similarity measure
\end{itemize}

\subsubsection{Control Experiments and Ablations}

To determine whether successful prediction reflects true
semantic encoding or methodological artefacts, we performed
the following ablations:

\begin{itemize}
    \item \textbf{Upper-bound experiments:} mapping a feature
    matrix to itself to estimate the methodological ceiling
    imposed by dataset sparsity. 

    \item \textbf{Random-feature baselines:} replacing feature
    values with random noise (\emph{Rand}) or random
    sparse feature assignments (\emph{Shuffle}). Surprisingly,
    these remain partly predictable due to sparsity-induced
    upper bounds. 

    \item \textbf{Taxonomic corruption:} replacing correct abstraction relations
    with random ones while leaving other features
    intact, to test sensitivity to core semantic structure.
    Performance does not meaningfully drop.

    \item \textbf{Structured nonsense features:} replacing
    continuous values by character-length differences
    (\emph{CDiff}), which nevertheless yield high Spearman
    correlations despite being semantically meaningless.

    \item \textbf{Neighborhood analysis:} testing whether the
    mapping preserves geometric similarity rather than feature
    structure, revealing that the method explains geometry,
    not semantic properties.
\end{itemize}

\begin{figure}
\centerline{\includegraphics[width=0.5\textwidth]{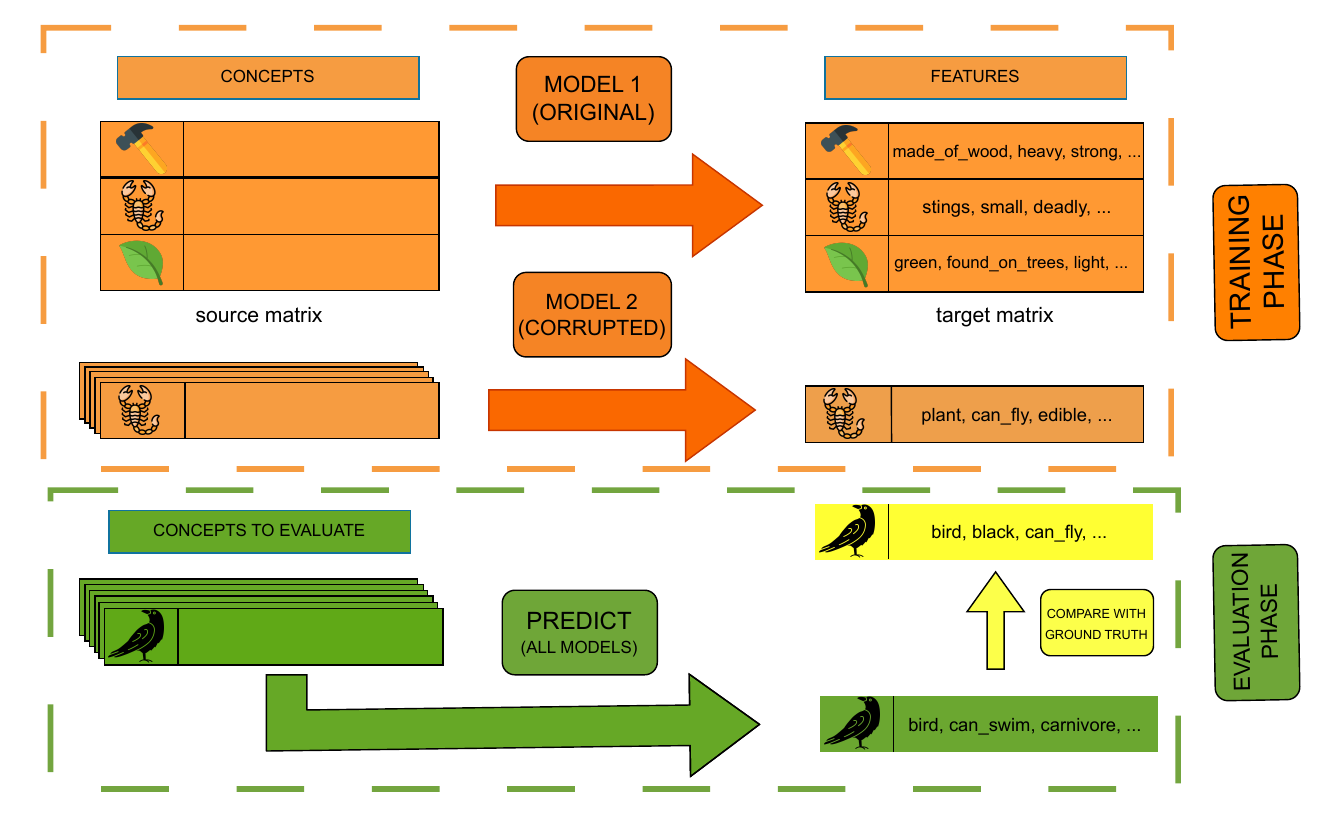}}
	 {\caption{Pipeline for mapping Embeddings. First the original feature norms are used, then random or nonsensical ones.}\label{fig:dataflow in b}}
\end{figure}
\section{Results and Discussion}

We summarize the empirical findings from the two
explanation pipelines. Details of the experimental setup are
provided in Section~3. Our goal is not to benchmark model
performance, but to report where and how the interpretability
assumptions underlying each method failed, despite correctly
implemented and validated pipelines.

\subsection{Attention-Based Pipeline}

\paragraph{Loss of token identity}
Residual-stream tracing showed that token representations
rapidly become mixtures of multiple upstream positions.
By mid-depth layers, cosine alignment between a token’s
hidden state and its original embedding dropped
substantially.

\paragraph{Attention maps remain structured even under perturbations}
Shuffling token embeddings, injecting noise into attention
logits, or swapping positions, produced attention visualizations
that still appeared linguistically structured.

Together, these results show that attention visualizations
maintain apparent structure even when the underlying token
representations no longer support a token-level explanation. The reasons here are the mechanisms of the attention module: First, the attention computations "melt" the embeddings of tokens into each other, so their representations mix before the layer output is generated. Second, there is no linear dependence between a layer's production and the input to the next layer of attention heads, because the feed-forward networks and layer normalization in between will change the vectors non-linearly. Despite those plain technical facts, token identity over all layers is commonly assumed.

\subsection{Embedding-Based Property Inference}

\textbf{a) Random and shuffled features remain highly predictable:}
Across all three feature-norm datasets, both PLSR and FFNN models
achieved substantial performance even when features were randomly
shuffled. As shown in Table~\ref{tab:prop-inference-upper-bounds},
shuffled feature matrices (preserving only sparsity and value ranges)
reached nearly the same scores as the original system mapping (Sys) and
approached the methodological upper bounds (Upper / Shuf-Upper). Fully
random dense features (Rand) performed poorly, indicating that the
predictability arises from structural properties of the norms rather
than semantic content. We used code and data from our previous work, both can be found on the Github Repository \cite{PredictionIsNotExplanation:Code}.

\begin{table}[t]
\centering
\caption{Results and methodological upper bounds for shuffled features and random baseline. Sys = original mapping from BERT to feature norms;
Upper = methodological upper bound (mapping norms to themselves).}
\begin{tabular}{lccccc}
\toprule
Norm & Sys & Upper & Shuffle & Shuf-Upper & Rand \\
\midrule
McRae (F1@10)    & 0.25 & 0.27 & 0.10 & 0.13 & 0.01 \\
Buchanan (F1@10) & 0.18 & 0.22 & 0.06 & 0.11 & 0.01 \\
Binder ($\rho$)  & 0.74 & 0.90 & 0.30 & 0.59 & 0.01 \\
\bottomrule
\end{tabular}
\label{tab:prop-inference-upper-bounds}
\end{table}
\textbf{b) Semantic corruption does not reduce performance:}
Replacing meaningful taxonomic hypernyms with random alternatives
produced no meaningful degradation in performance (e.g., McRae:
F1@10 decreases only from 0.25 to 0.23). The mapping models are thus insensitive to semantic identity and is dominated by geometric and sparsity-induced structure that cannot be distinguished from semantics by the method.

\textbf{c) Metrics reward geometric similarity, not semantic decoding:}
Neighborhood-accuracy analyses (Table~\ref{tab:prop-inference-na})
show that mapping methods primarily reproduce geometric cluster
structure. Even shuffled features achieve competitive NA@10 scores
and again approach their upper bounds, confirming that high predictive
scores reflect structural artifacts rather than semantic decoding.

\textbf{Summary:}
Across all evaluations, embedding-based property inference produced
convincing quantitative scores even when underlying assumptions were
violated. Models succeeded on shuffled, corrupted, and partially random
feature sets, and upper-bound analyses revealed that much of the
predictive success stems from intrinsic artifacts of the
datasets. Predictive performance alone therefore cannot be taken as
evidence that a model encodes or decodes semantic properties.

\begin{table}[t]
\centering
\caption{Neighborhood Accuracy (NA@10) for the original mapping (Sys) and the baselines, (up) are the upper bounds.}
\begin{tabular}{lcccccc}
\toprule
 & \multicolumn{2}{c}{Sys} & \multicolumn{2}{c}{Rand} & \multicolumn{2}{c}{Shuffle} \\
Norm & NA & Up & NA & Up & NA & Up \\
\midrule
McRae (NA@10)    & 0.37 & 0.49 & 0.19 & 0.22 & 0.19 & 0.23 \\
Buchanan (NA@10) & 0.18 & 0.28 & 0.02 & 0.04 & 0.02 & 0.07 \\
Binder (NA@10)   & 0.56 & 0.92 & 0.19 & 0.45 & 0.19 & 0.47 \\
\bottomrule
\end{tabular}
\label{tab:prop-inference-na}
\end{table}

\section{Lessons Learned}

Across both explanation pipelines, the experiments revealed
not only methodological failure points but also broader
lessons about how interpretability can mislead when its
assumptions remain untested. We summarize the key insights
below.

\noindent \textbf{a) Explanatory Capacity Must Be Tested, Not Assumed:}
Both attention-based analysis and property inference appeared
plausible when evaluated with standard tools. Only after
introducing diagnostic controls did their underlying premises
break down. This highlights the need for explicit assumption-testing as part of any interpretability workflow,
rather than treating explanatory claims as self-validating.

\noindent \textbf{b) Visual or Predictive Suggestion is not Understanding:}
Attention maps remained linguistically structured under
perturbations, and property inference achieved high scores
even on random or corrupted features. These outcomes
demonstrate that visual plausibility or predictive accuracy can mask fundamental misalignment between a method and the mechanisms it is meant to reveal.

\noindent \textbf{c) Dataset and Methodological Artifacts Dominate Results:}
Sparsity, low-rank constraints, geometric clustering, and other
structural properties of the models and datasets drove much of
the observed behavior. Without careful controls, these artifacts
can be mistaken for genuine evidence of semantic
understanding.

\noindent \textbf{d) Interpretability methods might fail deployment conditions:}
In pervasive computing scenarios, interpretability assumptions that appear valid in controlled settings may silently fail. Evaluating explanation methods without accounting for these conditions risks introducing false confidence into system design and maintenance. 

\noindent \textbf{e) Negative Results Are Essential for Interpretability:}
Finally, the failures documented here underscore the value of
negative results in interpretability research. Methods that seem
intuitive or convincing can fail in subtle ways, and identifying
these failure modes is necessary for building more reliable
explanation tools.

\section{Conclusion}
At the start of our work, we set out with a simple goal: take two of the most widely trusted approaches for explaining LLMs, namely attention analysis and embedding-based property inference, and use them to understand how models represent structure and meaning. What we found instead was that both methods were better at producing convincing stories than at producing actual explanations. 
\noindent In the attention experiments, we expected relational patterns to emerge cleanly: token A attends to token B, so the model must be representing that relation. What we found out is that residual mixing breaks the assumption that later-layer representations correspond to individual tokens. As a result, attention patterns cannot be treated as stable relational structures, even though they are often interpreted as such.
\noindent The embedding experiments failed differently, but with the same moral. Mapping embeddings to human-interpretable semantic features looked like a principled way to see what knowledge models encode. Yet as we ran more ablations, it became clearer that high predictive scores turned out to be driven by upper bounds and constraints of the feature norms rather than meaningful semantic content. Even corrupted or randomized feature structures yielded very high results. This undermines the assumption that successful prediction reflects encoded knowledge. 
\noindent Together, these findings demonstrate that both techniques provide outputs that are easy to interpret but difficult to explain scientifically. Failures of interpretability in pervasive and distributed systems affect how models are compressed, deployed, debugged, and trusted within real-world infrastructures. Our findings therefore highlight a class of system-level risks that arise when interpretability methods are treated as reliable explanations without testing their underlying assumptions. Our negative results therefore serve the constructive purpose of identifying methodological limits, clarifying which assumptions do not hold, and highlighting where more rigorous, mechanism-aware interpretability work is needed.

\bibliographystyle{IEEEtran}
\bibliography{IEEEfull}

\end{document}